\documentclass{article}

\usepackage{microtype}
\usepackage{graphicx}
\usepackage{subfigure}
\usepackage{booktabs} 

\usepackage{array}    
\usepackage{caption}  
\usepackage{float}

\usepackage[ruled,linesnumbered,noend]{algorithm2e}
\usepackage{algpseudocode}
\usepackage{hyperref}

\usepackage[accepted]{icml2025}

\usepackage{amsmath}
\usepackage{amssymb}
\usepackage{mathtools}
\usepackage{amsthm}
\usepackage{multicol}
\usepackage{multirow}

\usepackage[capitalize,noabbrev]{cleveref}

\theoremstyle{plain}

\theoremstyle{definition}

\theoremstyle{remark}

\usepackage[ruled,linesnumbered,noend]{algorithm2e}
\usepackage{xcolor}
\usepackage{colortbl}

\usepackage[textsize=tiny]{todonotes}

\icmltitlerunning{MMedPO: Aligning Medical Vision-Language Models with Clinical-Aware Multimodal Preference Optimization}

\begin{document}
\newcommand{\ours}{MMedPO}

\twocolumn[
\icmltitle{\ours: Aligning Medical Vision-Language Models with Clinical-Aware Multimodal Preference Optimization}




\icmlsetsymbol{equal}{*}

\begin{icmlauthorlist}
\icmlauthor{Kangyu Zhu}{equal,unc,brown}
\icmlauthor{Peng Xia}{equal,unc}
\icmlauthor{Yun Li}{unc}
\icmlauthor{Hongtu Zhu}{unc}
\icmlauthor{Sheng Wang}{uw}
\icmlauthor{Huaxiu Yao}{unc}
\end{icmlauthorlist}

\icmlaffiliation{unc}{UNC Chapel-Hill}
\icmlaffiliation{brown}{Brown University}
\icmlaffiliation{uw}{University of Washington}

\icmlcorrespondingauthor{Kangyu Zhu}{kangyu@unc.edu}
\icmlcorrespondingauthor{Peng Xia}{pxia@cs.unc.edu}
\icmlcorrespondingauthor{Huaxiu Yao}{huaxiu@cs.unc.edu}

\icmlkeywords{Machine Learning, ICML}

\vskip 0.3in
]

\author{
Kangyu Zhu$^{1,2}$\thanks{Work was done during Kangyu Zhu’s internship at UNC.},
Peng Xia$^{1}$\footnotemark[1],
Yun Li$^{1}$, Hongtu Zhu$^{1}$, Sheng Wang$^{3}$, Huaxiu Yao$^{1}$\\
$^{1}$UNC Chapel-Hill, $^{2}$Brown University,
$^{3}$University of Washington\\ \textit{kangyu@unc.edu, \{pxia,huaxiu\}@cs.unc.edu}
}



\printAffiliationsAndNotice{\icmlEqualContribution} 

\begin{abstract}
The advancement of Large Vision-Language Models (LVLMs) has propelled their application in medicine. However, Medical LVLMs (Med-LVLMs) encounter factuality issues due to modality misalignment, where the models prioritize textual knowledge over visual input, causing hallucinations that conflict with medical images. Previous attempts on preference optimization have inadequately mitigated clinical relevance in preference data, making these samples easily distinguishable and reducing alignment effectiveness.
To address this challenge, we propose \ours, a novel multimodal medical preference optimization approach that considers the clinical relevance of preference samples to enhance Med-LVLM alignment. \ours\ curates multimodal preference data by introducing two types of dispreference: (1) plausible hallucinations injected through target Med-LVLMs or GPT-4o to produce medically inaccurate responses, and (2) lesion region neglect achieved through local lesion-noising, disrupting visual understanding of critical areas. We then calculate clinical relevance for each sample based on scores from Med-LLMs and visual tools, and integrate these scores into the preference optimization process as weights, enabling effective alignment.
Our experiments demonstrate that \ours\ significantly enhances factual accuracy, achieving improvements over existing baseline methods by averaging 14.2\% and 51.7\% across the Med-VQA and report generation tasks. Our code are available in \href{https://github.com/aiming-lab/MMedPO}{https://github.com/aiming-lab/MMedPO}.
\end{abstract}

\section{Introduction}
\label{sec:intro}

Artificial intelligence is increasingly being applied in the medical field~\cite{tuauctan2021artificial,wang2019artificial,ye2021unified,tu2024towards,xia2024generalizing,wang2025screening,hu2024ophnet,hu2023nurvid,li2024tp}, including areas such as disease diagnosis and treatment planning. With the recent surge in popularity of Large Vision-Language Models (LVLMs)~\cite{liu2023visual,liu2023improved,zhu2023minigpt}, Medical LVLMs (Med-LVLMs) have begun to develop rapidly, drawing significant attention~\citep{li2023llava,moor2023med,zhang2023pmc,wu2023towards,xia2024rule,xia2024mmed}. However, these models still face the challenge of factuality, which is largely due to modality misalignment issues~\cite{cui2023holistic,zhou2024aligning,sun2024stllava}. Models with poor modality alignment tends to prioritize the textual knowledge learned during training over the actual visual input. As a result, Med-LVLMs often produce hallucinations, generating text that appears coherent but contradicts the information in the corresponding medical image~\citep{xia2024cares,royer2024multimedeval}. 

\begin{figure}[t]
    \centering
    \includegraphics[width=\linewidth]{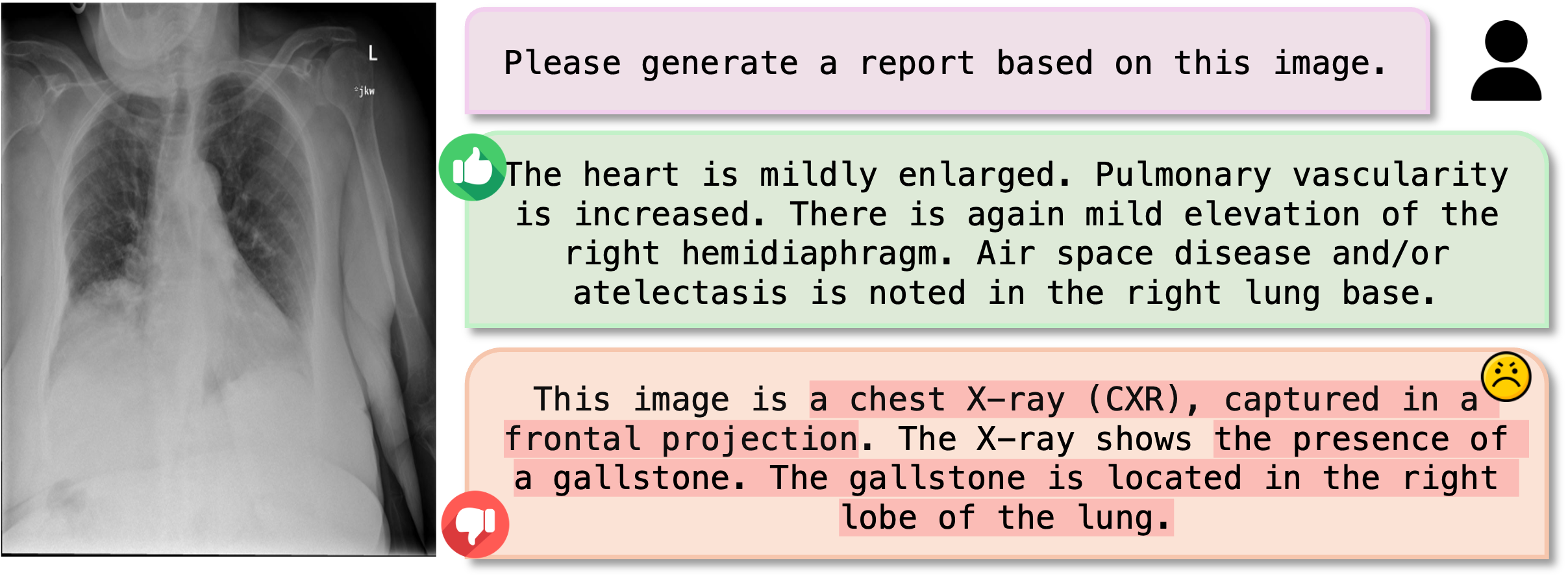}  
    \caption{An illustration of preference data pair. The dispreferred response contains nonfactual and clinically meaningless content.}
    \label{fig:low score exp}
\end{figure}

To tackle this issue, several studies have employed preference optimization on Med-LVLMs, aiming to improve alignment between medical image and text modalities with factuality improvement~\cite{hein2024preference, sun2024stllava, banerjee2024direct}. However, these methods simply leverage the preference data generation process used for aligning general LVLMs on natural images, overlooking the clinical relevance of the generated preference samples. Consequently, these preference samples become relatively easily distinguishable, reducing their effectiveness in aligning Med-LVLMs. Clinical relevance can be considered from two perspectives. First, in these preference samples, it is essential that both preferred and dispreferred responses are clinically meaningful; if dispreferred responses lack clinical relevance, Med-LVLMs can easily distinguish them, diminishing the sample’s effectiveness. For instance, a dispreferred response such as ``\textit{a gallstone in the right lobe of the lung...}" reflects a clear factual error with limited clinical relevance~\cite{tu2024towards}. Second, when improving alignment between the generated medical response and the input medical image, focused attention on local lesion areas is essential for accurate medical image understanding. Correcting dispreferred responses that arise from overlooking these lesion regions is crucial for achieving more precise medical alignment.

To address this challenge, we introduce \textbf{\ours}, a novel \textbf{M}ultimodal \textbf{Med}ical \textbf{P}reference \textbf{O}ptimization approach designed to quantify preference sample importance based on clinical relevance, enabling more effective preference optimization in Med-LVLMs. In \ours, we first curate multimodal medical preference data using two strategies: (1) introducing dispreference by leveraging target Med-LVLMs~\cite{li2023llava} or GPT-4o~\cite{openai2023gpt4} to inject plausible hallucinations into responses, ensuring dispreferred outputs contain evident medical inaccuracies, such as incorrect imaging interpretations, misleading descriptions, or inaccurate diagnoses; and (2) provoking dispreference by neglecting lesion regions through a visual tool-guided local lesion-noising process, which disrupts the model's understanding of these areas, leading to responses that overlook critical regions, thus being marked as dispreferred. We then quantify each preference sample's clinical significance by formulating sample importance scores, which integrate (1) clinical significance scores of dispreferred responses, evaluated by a multiple Med-LLMs collaboration process, and (2) confidence scores from visual tools to assess lesion region detection accuracy. These sample importance scores are then feed into a preference optimization process, enabling more effective alignment based on the clinical relevance of each preference sample.

The primary contribution of this paper is the introduction of \ours, aiming to quantify the clinical significance of curated preference samples to achieve more effective alignment and enhance factual accuracy in Med-LVLMs. Empirical results on two Medical Visual Question Answering (Med-VQA)~\cite{lau2018dataset,liu2021slake} and two report generation datasets~\cite{johnson2020mimic,demner2016preparing} demonstrate that \ours\ substantially improves the factual accuracy of Med-LVLMs, achieving significant gains over the best previous preference optimization methods, with improvements of 14.2\% and 51.7\% on the Med-VQA and report generation tasks, respectively.

\begin{figure*}[t]
  \centering
  \includegraphics[width=0.9\textwidth]{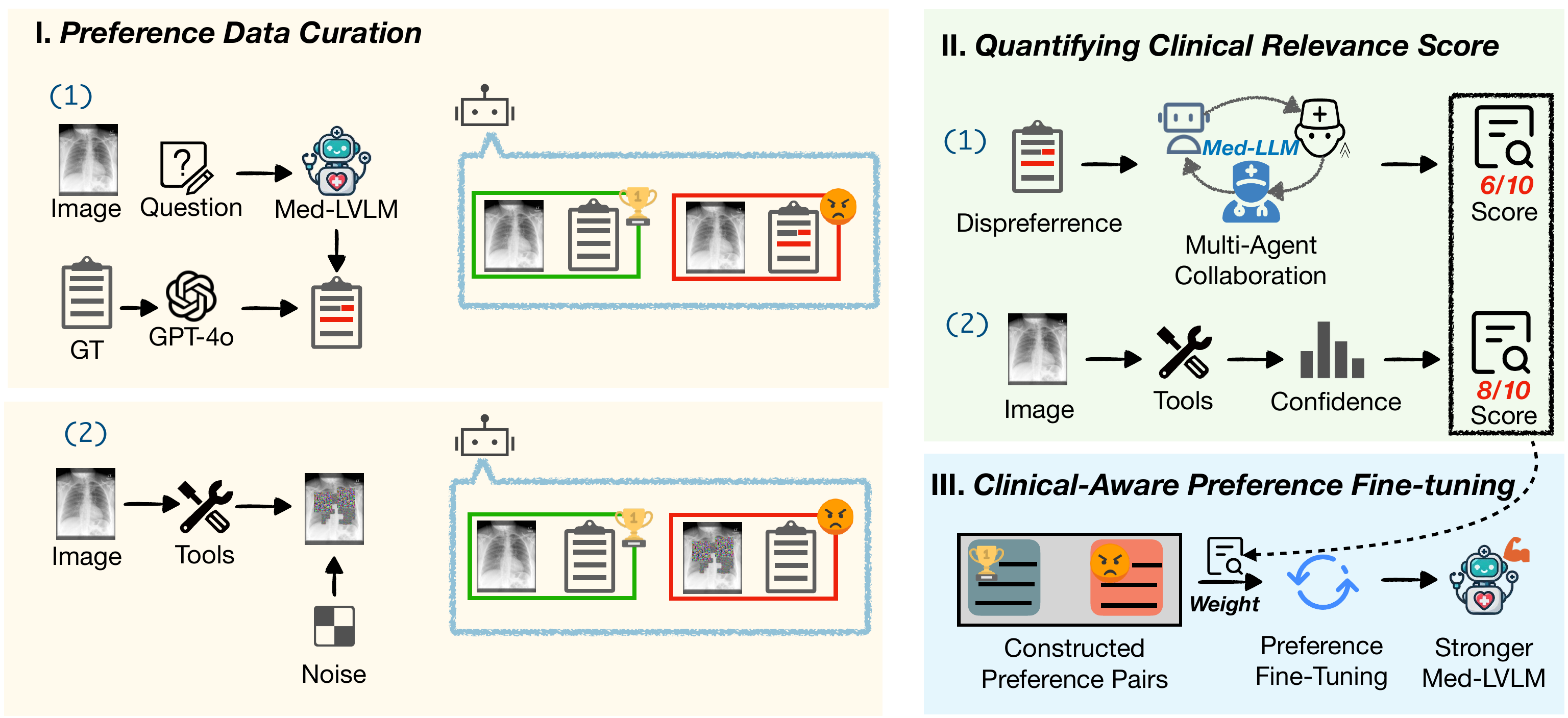}
  \vspace{-1em}
\caption{The overview of \ours\ outlines a comprehensive framework consisting of multimodal preference data curation, a quantified preference scoring module, and clinical-aware preference optimization. For data curation, the hallucinated text response and localized noisy images are joint constructed as preference data. Then the clinical relevance score is obtained through a multi-agent collaboration system and visual tools. Finally, these scores, serve as weights for the clinical-aware preference optimization.
}
  \label{fig:framework}
  \vspace{-1em}
\end{figure*}

\section{Preliminaries}

\subsection{Medical Large Vision Language Models}
Medical Large Vision-Language Models (Med-LVLMs) are advanced architectures primarily comprising a Large Language Model (LLM) integrated with a specialized visual module. The visual module analyzes medical images to extract relevant information, transforming it into a representation compatible with the LLM’s processing capabilities. Given a medical image $x_v$ and a clinical query $x_t$, the combined input is represented as $x = (x_v, x_t)$. The model then autoregressively predicts the probability distribution of the next token in the sequence, leveraging the multimodal input. The text output generated by the model is denoted as $y$. 

\subsection{Preference Optimization}
Preference optimization has proven highly effective in fine-tuning LLMs~\cite{rafailov2023direct,bai2022training}, leading to a significant alignment between model behavior and target objectives. In preference optimization, given an input $x$, the language model policy $\pi_\theta$ generates a conditional distribution $\pi_\theta(y \mid x)$, where $y$ represents the output text response. One of the notable methods, Direct Preference Optimization (DPO)~\cite{rafailov2023direct}, leverages preference data to facilitate alignment within LLMs. The preference dataset is defined as \begin{small}$\mathcal{D} = \{(x^{(i)}, y_w^{(i)}, y_l^{(i)})\}_{i=1}^{N}$\end{small}, where $y_w^{(i)}$ denotes the preferred response and $y_l^{(i)}$ the dispreferred response for a given input $x$. The probability of preferring $y_w$ over $y_l$ is modeled as $p(y_w \succ y_l) = \sigma(r(x, y_w) - r(x, y_l))$, with $\sigma(\cdot)$ representing the sigmoid function. In DPO, the optimization process is expressed as a following loss computed over the preference data:
\begin{equation}
\small
\begin{array}{l}
\mathcal{L}_{\textit{DPO}}(\pi_\theta; \pi_{\text{ref}}) = -\mathbb{E}_{(x,y_w,y_l) \sim \mathcal{D}} \\
\left[ \log \sigma
\left(
\alpha \log \frac{\pi_\theta(y_w | x)}{\pi_{\text{ref}}(y_w | x)}
- \alpha \log \frac{\pi_\theta(y_l | x)}{\pi_{\text{ref}}(y_l | x)}
\right) \right].
\end{array}
\label{eq:dpo}
\end{equation}
\vspace{-1em}

where $\pi_\theta$ represents the reference policy, which is the LLM fine-tuned through supervised fine-tuning.

\section{Multimodal Medical Preference Optimization (\ours)}
\label{sec:method}
In this section, we propose \ours, a clinical-aware multimodal preference optimization method to address modality misalignment challenges in Med-LVLMs, which consists of three steps and the entire framework is illustrated in Figure~\ref{fig:framework}. Firstly, we use the target Med-LVLM or GPT along with medical visual tools to jointly construct medical multimodal preference data. Second, we evaluate the clinical relevance of each preference sample using a collaborative process with multiple Med-LLMs and confidence scores from medical visual tools for lesion region detection. Lastly, the normalized clinical relevance scores are integrated into the preference optimization process to achieve clinical-aware preference optimization. We detail these steps as follows:

\subsection{Preference Data Curation}
\label{sec:data}

In the first step, our goal is to construct a high-quality, medical-specific multimodal preference dataset using two strategies: (1) introducing dispreference by using target Med-LVLMs or GPT-4o~\cite{openai2023gpt4} to inject hallucinations into medical responses, ensuring that dispreferred responses include significant medical inaccuracies; (2) provoking dispreference by neglecting lesion regions through a medical visual tool-augmented local lesion-noising process, resulting in dispreferred responses that overlook critical regions. We detail both strategies as follows:

\noindent \textbf{Generating Hallucinated Medical Responses}. In the first strategy, we aim to generate a hallucinated medical response, designated as the dispreferred response, while the ground truth serves as the preferred response. To achieve this, we first perform multiple rounds of sampling on the target Med-LVLMs $\mathcal{M}(\cdot)$ to collect a set of potential hallucinated responses. We then use GPT-4o to evaluate all candidate responses and select the response with the highest level of hallucination, displaying clear conflicts with the ground truth. If none of the candidates exhibit significant hallucinations, we use GPT-4o to generate a new hallucinated response based on ground truth to ensure that dispreference contain factual inaccuracies, such as incorrect imaging interpretations, misleading condition descriptions, or erroneous diagnoses. The preference pairs constructed using this strategy are denoted as \( \mathcal{D}_t \). 

\noindent \textbf{Adding Noise to Localized Lesion Region}. To improve the alignment between generated medical responses and input medical images, concentrated attention on localized lesion areas is vital for accurate interpretation. Thus, we construct dispreferred response that stem from neglecting these lesion regions. Specifically, we leverage a medical visual tool (e.g.,  MedKLIP~\cite{wu2023medklip}) $\mathcal{T}(\cdot)$, to predict disease-related local regions $h=\mathcal{T}(x_v)$ for each medical image $x_v$. We then introduce noise into these detected localized lesion regions within the original image. The noise step is defined as \( k \), and the noised image at step \( k \) can be expressed as follows: 

\vspace{-2em}
\begin{equation}
\small
\label{eq:noise}
x_v^* = \sqrt{\bar{\xi}_k} \cdot (x_v \odot h) + \sqrt{1 - \bar{\xi}_k} \cdot (\epsilon \odot h) + (x_v \odot (1 - h)),
\end{equation}
where \( \bar{\xi}_t = \prod_{i=0}^k \xi_i \) and \( \xi_k \in (0, 1) \) are hyperparameters. In this approach, the original image \( x_v \) paired with the ground truth \( y \) is considered preferred, while the image with localized noise \( x_k \) paired with the same ground truth \( y \) is regarded as dispreferred. The preference data constructed using this strategy is denoted as \( \mathcal{D}_v \). 

Finally, we merge the two preference sets generated by the above two strategies and denote the preference dataset as \begin{small}$\mathcal{D}_{o}=\mathcal{D}_{t} \cup \mathcal{D}_{v}=\{x^{(i)}, x^{*(i)}, y_{w}^{(i)}, y_{l}^{(i)}\}_{i=1}^N$\end{small}, where $x^{(i)}$ and $x^{*(i)}$ denote the normal and noisy input, \begin{small}$y_{w}^{(i)}$\end{small}, \begin{small}$y_{l}^{(i)}$\end{small} represent preferred and dispreferred responses, respectively.

\subsection{Quantified Clinical Relevance Score}
\label{sec:quatify}
After obtaining multimodal medical preference data, we will quantify the clinical relevance of each preference sample to drive effective optimization. Our hypothesis is that responses with higher clinical relevance are more valuable for preference optimization, while low-quality responses, in turn, reduce the effectiveness of optimization. We will explain in detail how clinical relevance is calculated below.

\subsubsection{Clinical Relevance Scores for Dispreferred Medical Responses}
For samples generated by the target Med-LVLM and GPT-4o (i.e., samples in $\mathcal{D}_t$), we evaluate the clinical relevance of the dispreferred response based solely on the model's internal medical knowledge, without the need for visual input~\cite{tian2024opportunities,thirunavukarasu2023large}. Including medical images for this evaluation is unnecessary and may even hinder the process. Therefore, we rely on Med-LLMs with high levels of medical expertise to assess the clinical relevance of these text responses. Moreover, relying on a single Med-LLM for evaluating clinical relevance may introduce bias and result in unreliable assessments~\cite{chanchateval}. To address this, we implement a multi-agent collaboration system comprising multiple Med-LLMs, each with varying levels of medical expertise. These Med-LLMs collaborate through a structured debating process to reach a consensus on clinical relevance scores, thereby improving the reliability of clinical relevance evaluations. 

Specifically, for each Med-LLM \( \mathcal{G}_i \), where \( 0 < i \leq g \) and \( g \) represents the total number of Med-LLMs, the objective of the multi-agent collaborative system is to establish consensus on the clinical relevance score across all agents (i.e., Med-LLMs). This process comprises $r$ rounds. In each round, each Med-LLM evaluates the clinical relevance score passed from the previous Med-LLM. The process begins with the first Med-LLM, \( \mathcal{G}_1 \), which evaluates a dispreferred response \( y_{l} \), generating a clinical relevance score \( s_1 = \mathcal{G}_1(y_{l}) \) and recording it in the score history \( S \). Subsequently, each following Med-LLM \( \mathcal{G}_{i} \) retrieves the prior scores \( s_{i-1} \) and determines whether to agree. If a Med-LLM concurs, it adopts \( s_{i-1} \) as its clinical relevance score \( s_{i} \); otherwise, it generates a new score as \( s_{i} \). This process continues until all Med-LLMs reach consensus and produce a final score. To prevent excessive evaluations, a threshold limits the number of evaluation rounds. If this threshold is reached before consensus, the final score is defined as the average of the scores in the history: $\hat{s} = \frac{\sum_{i=1}^{|S|} s_i}{|S|},$ ensuring efficient consensus that reflects clinical relevance, where \( |S| \) represents the total number of scores. 

\subsubsection{Confidence Scores for Localized Lesion Regions from Visual Tools}
For preference data in \( \mathcal{D}_v \), distinct noisy regions correspond to disease-related lesion areas. Introducing noise into images to generate dispreferred responses for preference comparison can improve the visual understanding of Med-LVLMs~\cite{zhou2024aligning,zhao2023beyond,wang2024mdpo}. Emphasizing lesions associated with the disease through noise can further enhance the model’s focus on these critical areas. However, if noisy regions are inaccurately defined, the reliability of these samples decreases, potentially impacting the model performance. Therefore, quantifying the accuracy of critical lesion detection to represent sample importance during optimization is importance. To achieve this, we use the confidence scores $s$ from visual tools that generate heatmaps of local regions as an indicator of clinical relevance. We assign different clinical relevance scores to preference pairs based on the confidence scores provided by visual tools for lesion detection.

\subsection{Clinical-Aware Preference Fine-tuning}
Following the previous steps, we construct multimodal medical preference data and assign a quantified clinical relevance score to each preference sample. During preference optimization, we treat this score as the sample weight representing the contribution of each preference data pair to the overall objective. To prevent underfitting caused by an excessively small overall loss, we apply a normalization strategy, mapping the scores to a fixed range while maintaining their mean and variance. Specifically, for each clinical relevance score \( s \), the normalized score \( s' \) is calculated as: \begin{small}$s' = \frac{(s - \mu)}{\sigma} $\end{small}, then we clip $s'$ to values of $[\alpha,\beta]$. Here $\alpha$ and $\beta$ denote the predefined upper and lower bounds for the normalized score, and $\mu$ and $\sigma$ represent the mean and variance of the original scores, respectively. After obtaining the normalized clinical relevance score, we fine-tune the Med-LVLM using a weighted DPO. Following Eqn.~\ref{eq:dpoo}, the adjusted loss with clinical relevance as sample weights is calculated as follows:
\begin{equation}
\begin{array}{l}
\mathcal{L}_{mmedpo} = -\mathbb{E}_{(x,x^*,y_{w},y_{l},s') \sim \mathcal{D}_{o}}  \\
\left[ s' \log \sigma
\left(
\alpha \log \frac{\pi_\theta(y_{w} | x)}{\pi_{o}(y_{w} | x)}
- \alpha \log \frac{\pi_\theta(y_{l} | x^*)}{\pi_{o}(y_{l} | x^*)}
\right) \right].
\end{array}
\label{eq:dpoo}
\end{equation}
\begin{algorithm}
\small
    \caption{Multimodal Medical Preference Optimization (\textbf{\ours})}
    \LinesNumbered
    \label{ag:dpo}
    \KwIn{$\mathcal{D}=\{x_v^{(i)},x_t^{(i)}, y^{(i)}\}_{i=1}^N$: Dataset; $\mathcal{M}(\cdot,\cdot)$: Med-LVLM; $\mathcal{T}(\cdot)$: Visual Tool; $\mathcal{G}(\cdot)$: Med-LLM; $\mathcal{N}(\cdot,\cdot)$: Localized Nosiy Process; $\mathcal{Z}(\cdot)$: Normalization.}
    
    \KwOut{$\pi_\theta$: Parameters of the Med-LVLM.}
    
    Initialize $\mathcal{D}_o$ with an empty set\\
    \ForEach{$(x_v, x_t, y) \in \mathcal{D}$}{
    $\triangleright$ \textit{\textcolor{blue}{Preference Data Curation}} \\
        Generate responses of the Med-LVLM 
        $a \leftarrow \mathcal{M}(x_v, x_t)$ \\
        Select the dispreferred response 
        $y_{l} \leftarrow \text{GPT}(a, y)$ \\
        $\triangleright$ \textit{\textcolor{blue}{Quantify the Clinical Relevance}} \\
        Quatify the clinical relevance using Med-LLMs
        $s_t \leftarrow \mathcal{G}(y_{l})$ \\
        Put $\{x_v, y, y_{l}, s_t\}$ into $\mathcal{D}_o$\;
        Obtain the heatmap of lesion region 
        $h \leftarrow \mathcal{T}(x_v)$ \\
        Save the confidence score from visual tool
        $s_v \leftarrow P(h|x_v)$ \\
        Add noise to the localized region
        $x_v^* \leftarrow \mathcal{N}(x_v, h)$ \\
        Put $\{x_v, x_v^*, y, s_v\}$ into $\mathcal{D}_o$\;
    }
    $\triangleright$ \textit{\textcolor{blue}{Clinical Preference Optimization}} \\
    \ForEach{$(x, x^*, y_{w}, y_{l}, s) \in \mathcal{D}_o$}{
        Normalize the score $s' \leftarrow \mathcal{Z}(s)$ \\
        Update $\pi_\theta$ through Eq.~\eqref{eq:dpoo} \\
    }
    
\end{algorithm}

\begin{table*}[htbp]
\centering
\footnotesize
\caption{Performance comparison on medical VQA and report generation tasks covering SLAKE, VQA-RAD, and IU-Xray datasets. For open-ended questions, we report recall (Open); for closed-ended questions, accuracy (Closed). The BLEU score denotes the average of BLEU-1/2/3/4. +SFT indicates that the model is first fine-tuned with SFT before applying the corresponding baselines. The best results and second best results are highlighted in \colorbox{red!25}{red} and \colorbox{blue!15}{blue}, respectively.}
\vspace{0.5em}
\resizebox{\linewidth}{!}{
\begin{tabular}{l|cc|cc|ccc|ccc}
\toprule
\multirow{2}{*}{Models} & \multicolumn{2}{c|}{\textbf{SLAKE}} & \multicolumn{2}{c|}{\textbf{VQA-RAD}} & \multicolumn{3}{c|}{\textbf{IU-Xray}} & \multicolumn{3}{c}{\textbf{MIMIC-CXR}} \\ 
 & Open & Closed & Open  & Closed & BLEU  & ROUGE-L & METEOR  & BLEU  & ROUGE-L & METEOR  \\ \midrule
LLaVA-Med v1.5          & 44.26  & 61.30         & 29.24  & 63.97         & 14.56 & 10.31 & 10.95 & 10.25 &9.38 &7.71\\ \midrule
+ Self-Rewarding & 42.63 &  61.30 & \cellcolor{blue!15}{33.29} & 64.17 & 14.20 & 10.38 & 10.52 & 10.78 & 9.27 & 7.73 \\
+ DPO   & 49.30   & 62.02        & 29.76  & 64.70   & 16.08 & 12.95 & 17.13 & 11.19 & 9.45 & 7.80 \\ 
+ POVID & 52.43 & 70.35 & 31.77 & \cellcolor{blue!15}{65.07} & 20.80 & 24.33 & 30.05 & 11.21 & 9.66 & \cellcolor{blue!15}{7.84}\\
+ SIMA & 51.77 & 69.10 & 31.23 & 64.80 & 17.11 & 22.87 & 29.10 & 11.16 & 9.58 & 7.49 \\
+ FiSAO & \cellcolor{blue!15}{52.69} & \cellcolor{blue!15}{70.46} & 32.70 & 64.11 & \cellcolor{blue!15}{21.06} & \cellcolor{blue!15}{25.72} & \cellcolor{blue!15}{30.82} & \cellcolor{blue!15}{11.32} & \cellcolor{blue!15}{9.68} & 7.62 \\
+ STLLaVA-Med & 48.65 & 61.75 & 30.17 & 64.38 & 16.11 & 10.58 & 10.51 & 11.11 & 9.29 & 7.72 \\ 
+ \textbf{\ours (Ours)}  
& \cellcolor{red!25}{53.99} 
& \cellcolor{red!25}{73.08}  
& \cellcolor{red!25}{36.36}  
& \cellcolor{red!25}{66.54}   
& \cellcolor{red!25}{23.49}      
& \cellcolor{red!25}{29.52}       
& \cellcolor{red!25}{34.16}    
& \cellcolor{red!25}{12.85} 
&\cellcolor{red!25}{11.13} 
& \cellcolor{red!25}{10.03} \\
\midrule
+ SFT & 50.45 & 65.62 & 31.38 & 64.26 & 22.75& 28.86 & 33.66 & 12.39 & 10.21 & 8.75 \\
\quad + Self-Rewarding & 50.62 & 65.89 & 32.69 & \cellcolor{blue!15}{65.89} & 22.89 & 28.97 & 33.93 & 12.15 & 10.05 & 8.77 \\
\quad + DPO  & \cellcolor{blue!15}{53.50} & 69.47 & 32.88 & 64.33 & 23.07 & \cellcolor{blue!15}{29.97} & 34.89 & 12.37 & 10.38 & 9.10 \\
\quad + POVID &  52.18 & 70.67 & 32.95 & 64.97 &\cellcolor{blue!15}{23.95} & 29.75 & 34.63 & 11.85 &10.45 & 9.05\\
\quad + SIMA & 51.75 & 69.28 & 32.50 & 64.08 &23.90 & 29.41 & 34.45 & 12.44 & 10.25 & 9.02 \\
\quad + FiSAO & 52.80 & \cellcolor{blue!15}{70.82} & 32.94 & 65.77 & 23.57 & 29.88 & \cellcolor{blue!15}{35.01} & \cellcolor{blue!15}{12.97} & \cellcolor{blue!15}{10.69} & \cellcolor{blue!15}{9.39} \\
\quad + STLLaVA-Med & 52.72 & 66.69 & \cellcolor{blue!15}{33.72} & 64.70 & 22.79 & 28.98 & 34.05 & 12.21 & 10.12 & 8.98 \\
\quad + \textbf{\ours (Ours)}  & \cellcolor{red!25}{55.23} & \cellcolor{red!25}{75.24} & \cellcolor{red!25}{34.03} & \cellcolor{red!25}{67.64} & \cellcolor{red!25}{24.00} & \cellcolor{red!25}{30.13} &\cellcolor{red!25}{35.17} &\cellcolor{red!25}{13.28} & \cellcolor{red!25}{13.22} & \cellcolor{red!25}{10.20} \\
\bottomrule 
\end{tabular}
}
\vspace{-1em}
\label{tab:results}
\end{table*}

\section{Experiment}
\label{sec:exper}

In this section, we evaluate the effectiveness of \ours\ to answer the following questions: (1) Can \ours\ enhance the factual accuracy of Med-LVLMs compared to other alignment baselines? (2) How does each individual component of the framework contribute to overall performance? (3) Can \ours\ be compatible with different Med-LVLM architectures? (4) Does \ours\ improve Med-LVLMs’ responses in terms of clinical relevance?
\subsection{Experimental Setups}

\noindent\textbf{Evaluation Datasets.}
To verify the effectiveness of \ours\ in improving factuality, we utilize four medical datasets: two medical VQA datasets, i.e., VQA-RAD~\cite{lau2018dataset} and SLAKE~\cite{liu2021slake}, and two report generation datasets, i.e., MIMIC-CXR~\cite{johnson2020mimic} and IU-Xray~\cite{demner2016preparing}. 

\noindent
\textbf{Implementation Details.}
We utilize LLaVA-Med-1.5 7B~\cite{li2023llava} as the base model. During the preference optimization stage, we apply LoRA fine-tuning~\cite{hu2021lora}, with a batch size of 4, a learning rate of 1e-7, and train for 3 epochs. For curating preference data, we use GPT-4o to evaluate and generate dispreferred responses. In the multi-agent collaboration system, multiple Med-LLMs, including LLaMA3-Med42-7B~\cite{christophe2024med42}, LLaMA3-Med42-70B, BioMistral-7B~\cite{labrak2024biomistral}, are used to evaluate the relevance scores for the preference data. See Appendix~\ref{sec:setting} for more details.

\noindent
\textbf{Baselines.}
We compare \ours\ with Direct Preference Optimization (DPO)~\cite{rafailov2023direct} and its variants, including the self-rewarding method~\cite{yuanself} and STLLaVA-Med~\cite{sun2024stllava}. In the self-rewarding method, the model generates its own responses to form preference pairs, while STLLaVA-Med further refines the preference selection process using GPT-4o and apply it in Med-LVLMs. We further compare three VLM preference fine-tuning methods originally designed for natural images: POVID~\cite{zhou2024aligning}, FiSAO~\cite{cui2024fine}, and SIMA~\cite{wang2024enhancing}. Additionally, we evaluate \ours\ and all baselines on models that have undergone supervised fine-tuning (SFT) with the corresponding datasets and compare their performance. 
Please see more details in Appendix~\ref{sec:baseline}.


\noindent
\textbf{Evaluation Metrics.}
For Med-VQA task, we use accuracy and recall for both closed-ended and open-ended questions. For the report generation task, we use BLEU Score~\cite{papineni2002bleu}, ROUGE-L~\cite{lin2004rouge} and METEOR~\cite{banerjee2005meteor} as the metrics.

\subsection{Main Results}
In this section, we present a comprehensive comparison of \ours\ with baseline methods. 

\noindent\textbf{Comparison with Baseline Methods.}
As shown in Table \ref{tab:results}, we evaluate our model’s performance against the original LLaVA-Med-1.5 and several preference optimization baselines. \ours\ demonstrates superior performance across both Medical VQA and report generation tasks. Specifically, for Med-VQA task, \ours\ significantly outperforms the best baseline (i.e., original DPO) by an average of 15.8\% and 10.3\% across the open-ended and closed-ended questions, respectively. We also observe that the overall performance improvement on open-ended questions is greater than that on closed-ended questions, indicating that \ours\ is particularly effective in guiding accurate open-ended generation. Additionally, \ours\ exhibits superior performance on the report generation task, surpassing the best baseline by 61.9\% and 26.0\% on IU-Xray and MIMIC-CXR, respectively. This demonstrates that, by constructing a multimodal preference dataset and assigning quantified clinical relevance scores to measure sample importance, \ours\ ensures that clinical relevance is fully considered during the preference optimization process, resulting in more accurate and clinically meaningful responses. 

\noindent \textbf{Comparison with Baseline Methods Enhanced by SFT.}
To demonstrate the compatibility of our approach with other training methods, we conduct experiments by applying \ours\ and other baseline methods to SFT. As shown in Table~\ref{tab:results}, \ours\ consistently outperforms the SFT baseline across all four datasets, with an average improvement of 14.2\%. When compared to other baselines applied to SFT, \ours\ achieves significantly better performance, with an average improvement of 10.5\%. These results further corroborate the effectiveness and compatibility of our approach, demonstrating its ability to integrate seamlessly with other training techniques to enhance model alignment.

\subsection{Quantitative Analysis}
In this section, we first conduct ablation study to analyze the effectiveness of each strategy and component in \ours\ for enhancing factual accuracy. Then, we evaluate the model’s compatibility with different backbones. We further explore how our approach improves Med-LVLMs’ responses in terms of clinical significance and visual understanding.

\subsubsection{Ablation Study}
\begin{figure}[h]
    \centering
    \includegraphics[width=\linewidth]{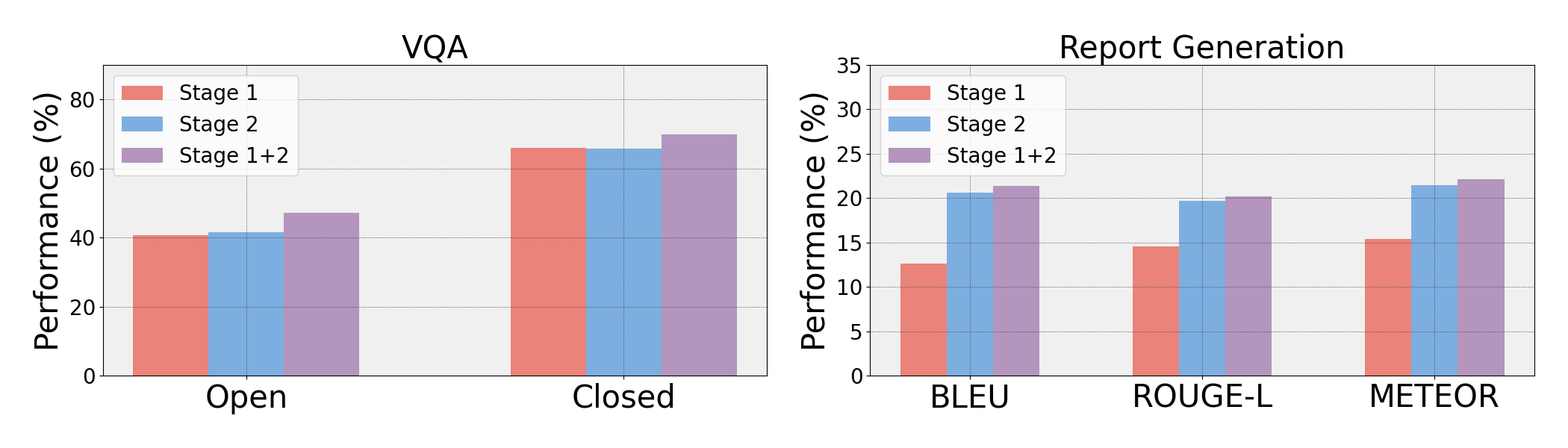}  
    \vspace{-1em}
    \caption{Comparison of the effectiveness of different preference curation strategies. ``stage 1": generating hallucinated medical responses; ``stage 2": adding noise to localized lesion regions; ``stage 1+2": merged  preference data. We report the average score on each dataset.
    }
    \label{fig:weight_distribution}
    \vspace{-1em}
\end{figure}

\noindent \textbf{Different Preference Curation Strategies. } To assess the impact of different preference curation strategies in \ours, namely generating hallucinated medical responses and adding noise to localized lesion regions, we evaluated their performance on these two components. The results, presented in Figure~\ref{fig:weight_distribution}, reveal that adding noise to localized lesion regions has a more pronounced effect on open-ended generation tasks (e.g., report generation) compared to generating hallucinated medical responses. For medical VQA tasks, the performance improvements from both preference curation processes are comparable. By integrating both strategies, \ours\ achieves the best overall performance across four datasets, effectively combining their strengths to maximize performance gains.

\begin{table}[htbp]
\centering
\caption{Comparison of performance across different datasets with and without clinical relevance score (CRS) for different preference curation strategies. Here, stage 1 and stage 2 denote generating hallucinated medical responses and adding noise to localized lesion regions, respectively. We report the average score on each dataset.}
\resizebox{\linewidth}{!}{
\begin{tabular}{l|c|c|c|c}
\toprule
 & \textbf{SLAKE} & \textbf{VQA-RAD} & \textbf{IU-Xray} & \textbf{MIMIC-CXR} \\
\midrule
Stage 1 w/o CRS & 55.65 & 47.23 & 10.95 & 6.55 \\
Stage 1 w CRS & \textbf{57.62} & \textbf{48.67} & \textbf{15.66} & \textbf{6.58} \\ \midrule
Stage 2 w/o CRS & 60.59 & 45.94 & 19.30 & 7.17 \\
Stage 2 w CRS & \textbf{60.88} & \textbf{46.97} & \textbf{25.00} & \textbf{7.24} \\
\bottomrule
\end{tabular}
}
\label{tab:aba}
\vspace{-1.5em}
\end{table}

\noindent\textbf{Clinical Relevance Score.}
To investigate the role of clinical relevance score as weight in the preference optimization process, we compare the results of applying this weight versus not applying it under different preference curation strategies. The results indicate that incorporating clinical relevance scores as weights in preference optimization improves the effectiveness of fine-tuning. Specifically, as shown in Table~\ref{tab:aba}, for VQA task, models utilizing clinical relevance scores as weights consistently outperform those without them, with an average improvement of 2.3\%. Also, significant performance gains are observed on the report generation task, where clinical relevance scores contributed positively across different preference curation strategies, achieving a clear average margin of 18.5\%. The clinical relevance scores assigned to each preference pair provide positive benefits to preference optimization, helping the Med-LVLMs generate responses that are more clinically meaningful and accurate.

\subsubsection{Multiple vs. Single Med-LLM}
To explore the impact of the multi-agent collaboration mechanism in generating clinical relevance scores, we conduct analytical experiments, comparing the performance using clinical relevance scores from single Med-LLM and multiple Med-LLMs. As shown in Table~\ref{tab:average_results_all_datasets}, we find that the consensus scores reached by multiple Med-LLMs positively contribute to performance improvement by an average of 3.6\% over four datasets. This aligns with our expectations, as relying on a single Med-LLM will introduce biases. The observed improvement is driven by reduced bias through the collaborative efforts of multiple Med-LLMs, resulting in more accurate and clinically meaningful relevance evaluations. In addition, the performance gains on the Med-VQA task using multiple Med-LLMs are notably larger compared to the report generation task. This may be attributed to greater disagreement among Med-LLMs on rejected VQA answers, allowing them to benefit more from achieving consensus. 

\subsubsection{Impact of Localized Lesion Noise}
To evaluate the impact of localized lesion noise during the preference optimization process, we compare the performance of preference data composed of images with localized noise versus those with global noise. Global noise refers to adding noise uniformly across the entire image. As shown in Table~\ref{tab:noise_comparison}, introducing localized noise consistently outperforms global noise across the four datasets. This indicates that lesion regions detected by visual tools are more prominent than the entire image. Introducing localized noise based on these regions helps the model better understand critical lesions, leading to more factually accurate responses.

\begin{table}[t]
\centering
\footnotesize
\caption{Comparison of model performance using clinical relevance scores from single Med-LLM and multiple Med-LLMs for \ours. We report the average score on each dataset.}
\resizebox{\linewidth}{!}{
\begin{tabular}{l|c|c|c|c}
\toprule
Models& \textbf{SLAKE} & \textbf{VQA-RAD} & \textbf{IU-Xray} & \textbf{MIMIC-CXR} \\ 
 \midrule
Single-LLM & 56.09 & 48.67 & 15.67 & 6.58 \\
Multi-LLMs & \textbf{57.53} & \textbf{51.14} & \textbf{15.86} & \textbf{6.86} \\ \bottomrule
\end{tabular}
}
\label{tab:average_results_all_datasets}
\vspace{-1.5em}
\end{table}


\begin{table}[t]
\centering
\footnotesize
\caption{Performance comparison between introducing local noise and global noise on the stage of constructing preference data by adding noise to medical images. 
}
\resizebox{\linewidth}{!}{
\begin{tabular}{l|c|c|c|c}
\toprule
Noise Location & \textbf{SLAKE} & \textbf{VQA-RAD} & \textbf{IU-Xray} & \textbf{MIMIC-CXR} \\
\midrule
Global &58.88 & 46.91 & 24.88 & 6.80 \\
Local & \textbf{59.88} & \textbf{46.98} & \textbf{25.00} & \textbf{7.24} \\
\bottomrule
\end{tabular}
}
\label{tab:noise_comparison}
\vspace{-1.5em}
\end{table}

\subsubsection{Compatibility Analysis}
\begin{figure}[htbp]
    \centering
    \includegraphics[width=\linewidth]{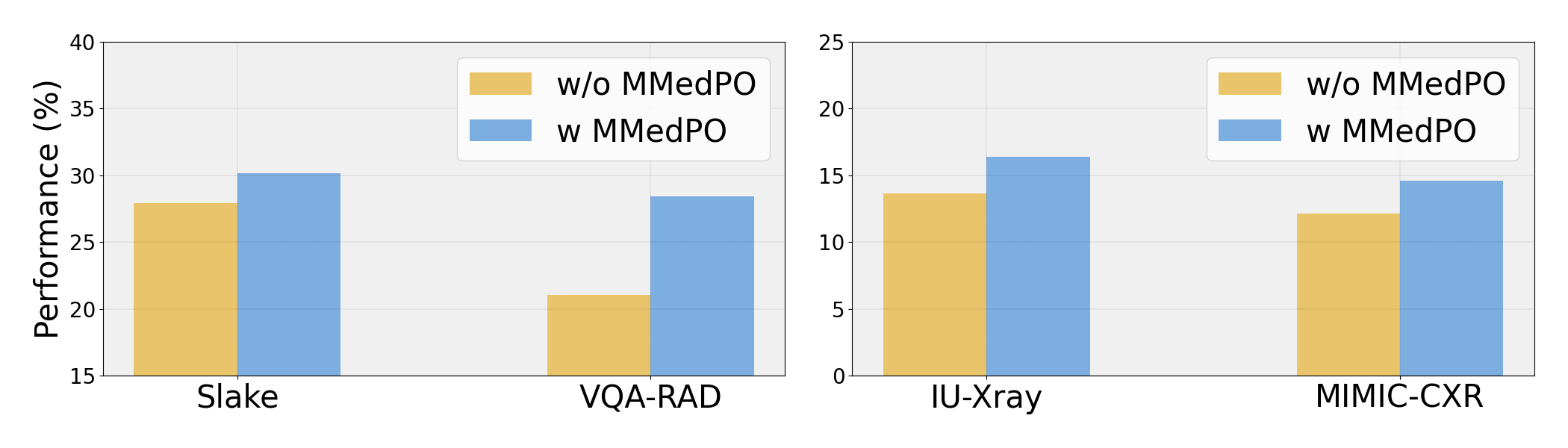}  
    \vspace{-1em}
    \caption{Analysis of compatibility using LLaVA-Med++ as the backbone model. Averaged metrics across datasets are presented.}
    \vspace{-1em}
    \label{fig:backbone_analysis}
\end{figure}

To evaluate the compatibility of our approach with different base models, particularly more powerful backbone architectures, we replace the backbone of LLaVA-Med-1.5 and conduct a series of experiments based on this configuration. Specifically, we apply our method to LLaVA-Med++~\cite{xie2024medtrinity}, which uses LLaMA-3~\cite{dubey2024llama} as language backbone and enhances its performance using a large-scale medical multimodal dataset MedTrinity-25M. As illustrated in Table~\ref{fig:backbone_analysis}, similar to the results obtain with LLaVA-Med-1.5, applying \ours\ leads to performance improvements across all four datasets. These findings highlight the strong compatibility and effectiveness of our approach when integrated with other powerful Med-LVLMs. \ours\ can be transferred to a wider range of base models, demonstrating strong generalizability for applications in clinical scenarios.

\subsection{Qualitative Analysis and Case Study}
In this section, we further conduct qualitative experiments and case analyses.

\subsubsection{Qualitative Analysis}
\begin{figure}[h]
    \centering
    \includegraphics[width=\linewidth]{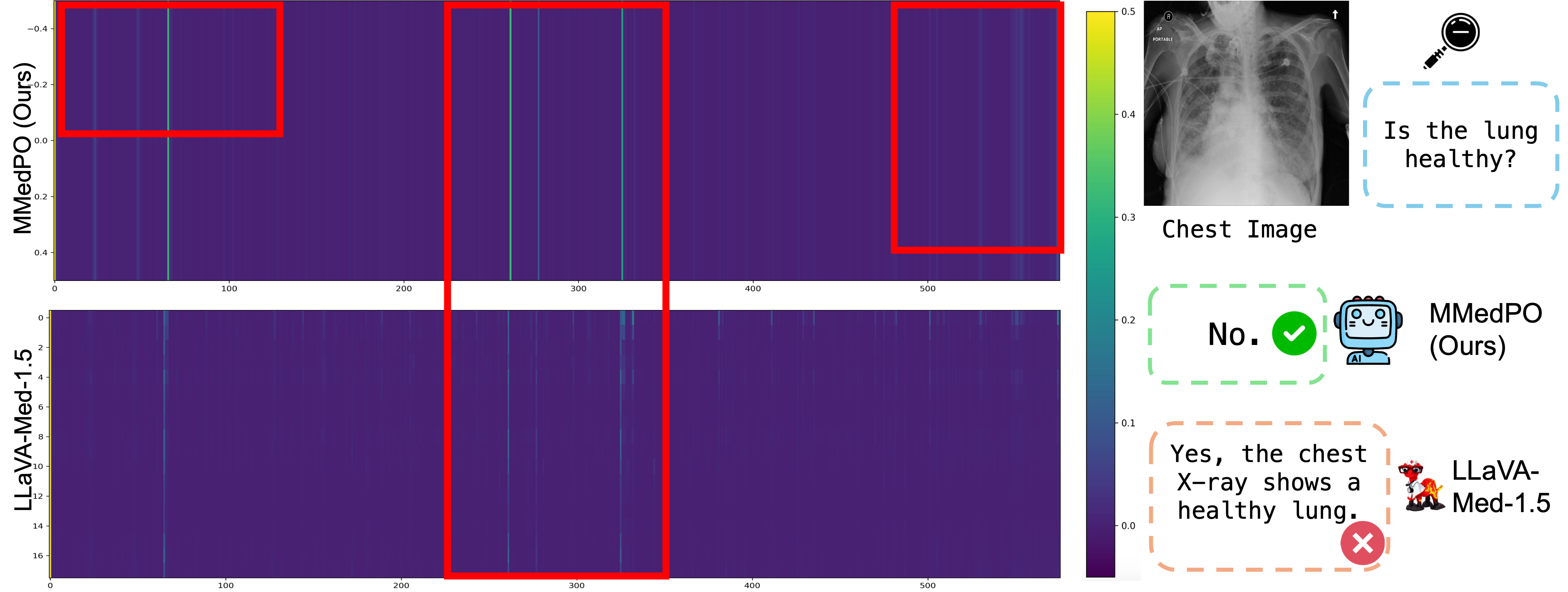}  
    \caption{Visualization of attention map of image tokens. The red box region is labeled with the attentions that are enhanced.}
    \label{fig:vis}
\end{figure}

\noindent\textbf{How does \ours\ in Improving Visual Understanding?}
To better understand the model’s visual comprehension capability, we visualize its attention map on image tokens. As shown in Figure~\ref{fig:vis}, compared to the attention map of the original model, the utilization of \ours\ significantly enhances the model's focus on visual information, particularly on critical lesion areas. This allows the model to extract sufficient information from visual inputs and improve consistency between text and images. Thus the model can reduce hallucinations and provide more accurate answers.

\begin{figure*}[t]
  \centering
  \includegraphics[width=0.85\textwidth]{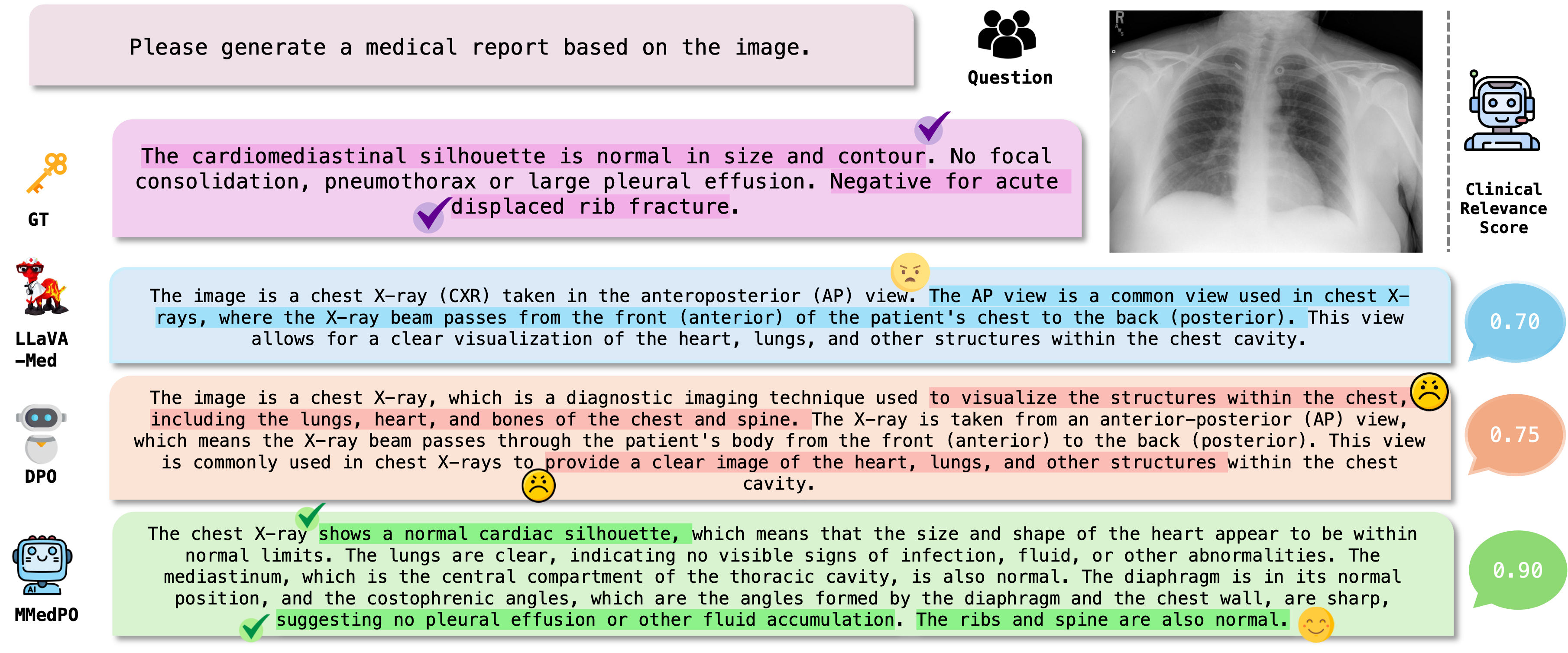}  
\vspace{-1em}
\caption{Examples demonstrating the clinical relevance of responses generated by \ours. Our approach not only enhances the factual accuracy but also significantly improves the clinical relevance, including various meaningful medical-level explanations.}
  \label{fig:sig}
  \vspace{-1em}
\end{figure*}

\noindent\textbf{Analysis Clinical Significance of Model's Response.}
Through the analysis of previous results, Med-LVLMs enhanced by \ours\ demonstrate a significant improvement in factuality accuracy. Additionally, from the clinical perspective, we aim to evaluate the clinical significance of the responses to verify the effectiveness of \ours\ in enhancing the clinical relevance of the model's outputs. As demonstrated in Figure~\ref{fig:sig}, Med-LVLMs with \ours\ outperforms both the original model and the one applied with DPO. The response with \ours\ accurately capture the condition of the cardiac silhouette and rib fracture in the image, aligning with the ground truth. This also improves clinical significance judged by Med-LLMs, whereas the original model and other baselines produced duplicate and clinically irrelevant content. The evaluation of response using clinical relevance from Med-LLMs quantitatively shows that \ours\ consistently achieves significantly higher clinical relevance scores. 

\begin{figure}[h]
    \centering
    \includegraphics[width=0.9\linewidth]{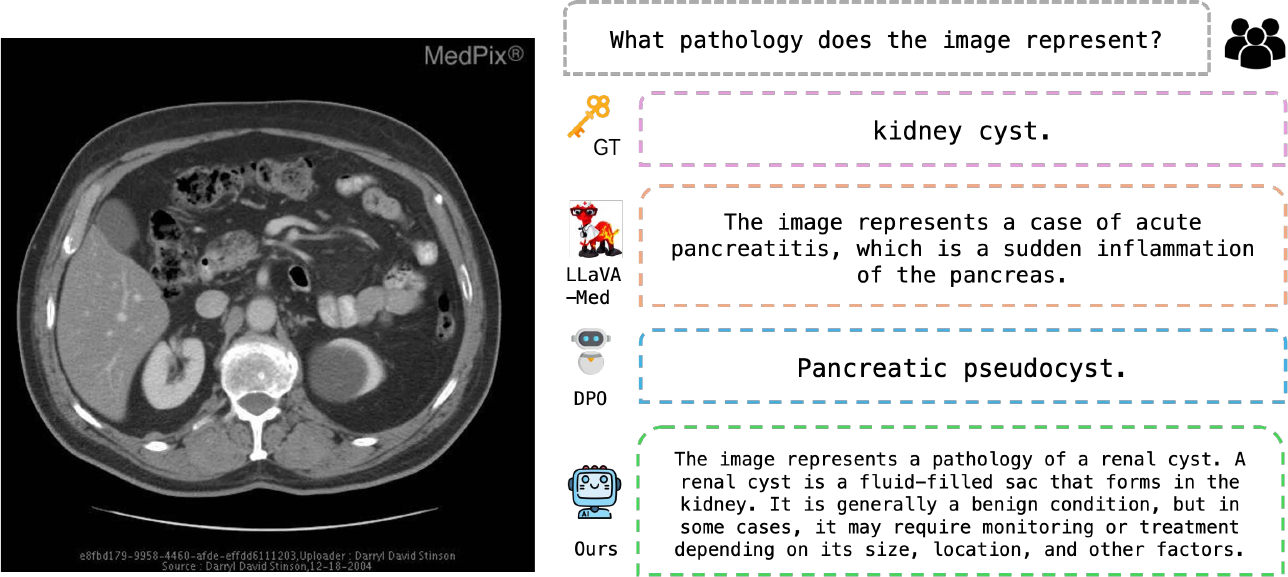} 
    \vspace{-1em}
    \caption{Illustration of factuality enhancement by \ours.}
    \label{fig:VQA case}
    \vspace{-1em}
\end{figure}
\subsubsection{Case Study}
We analyze two examples from Medical VQA task to illustrate how the model fine-tuned with \ours\ reduces factuality errors. As illustrated in Figure~\ref{fig:VQA case}, \ours\ shows improved performance in factual accuracy. In this case, when asked about pathology, \ours\ provides a more detailed response, focusing on the problem of cyst, which is similar to the ground truth, outperforming both LLaVA-Med and LLaVA-Med with DPO. This demonstrates that \ours\ effectively reduces hallucinations in Med-LVLMs, minimizing factual errors in multimodal understanding tasks.

\section{Related Work}
\textbf{Factuality Issues in Med-LVLMs}.
The development of Large Vision-Language Models (LVLMs) is progressing rapidly~\cite{liu2023improved,liu2023visual,zhu2023minigpt,bai2023qwenvl,xia2025hgclip,xia2024lmpt,han2025mdocagent,xia2024mmie}, which has, in turn, driven advancements in Medical Vision-Language Models (Med-LVLMs), achieving promising results in the medical field~\cite{li2023llava,moor2023med,thawkar2023xraygpt,wu2023towards}. However, the current Med-LVLMs still exhibit significant factual errors~\cite{wu2023can,li2023comprehensive,xia2024cares,chen2024detecting,jiang2024medthink,su2024conflictbank}. For example, they often lack sufficient judgment ability for complex content, and frequently generate responses with hallucinations that contradicts the visual information provided. This issue is particularly pronounced in medical domain, as it can potentially lead to misdiagnoses or missed diagnoses. Recently, there are several benchmarks~\cite{xia2024cares,royer2024multimedeval} that highlight the factuality issues of Med-LVLMs on multiple tasks such as the visual question answering and report generation.

\noindent
\textbf{Preference Optimization in Med-LVLMs.}
Aligning with human preferences for large models is an effective way to address hallucination issues~\cite{lee2023rlaif,zhou2024aligning,zhou2024calibrated,deng2024enhancing}. Preference fine-tuning in LVLMs generally involves two approaches: one aligns models based on human feedback~\cite{bai2022training,rafailov2023direct}, while the other uses feedback generated by AI~\cite{lee2023rlaif,zhou2024aligning,zhou2024calibrated,wang2024mdpo,zhou2025anyprefer,tong2025mj,su2024timo}. Recently, the preference fine-tuning technique has also been adapted for medical imaging~\cite{banerjee2024direct,sun2024stllava,hein2024preference} by generating dispreferred responses using GPT-4 or the target Med-LVLM. Although these methods have shown promise, they neglect the clinical relevance of generated samples. In Med-LVLMs, local visual information is crucial for accurate responses, yet current approaches rarely guide the model’s focus to specific lesion areas during preference fine-tuning. To tackle these issues, we incorporate quantified clinical relevance scores as weights to enhance modality alignment and introduce localized noise in medical images to construct dispreference, improving its understanding of key lesions.

\section{Conclusion}
\label{sec:conclusion}
In this work, we propose a novel clinical-aware multimodal preference optimization approach named \ours\, which considers the clinical relevance of each preference sample in preference optimization process. This method enhances Med-LVLM alignment while effectively reducing factual hallucinations. Specifically, to construct multimodal preference data, we introduce plausible hallucinations and apply local noise to critical lesion regions. Furthermore, we assign clinical relevance for data samples through Med-LLMs and visual tools, and then incorporate these scores as weights in the preference fine-tuning process. We evaluate the effectiveness of \ours\ on the Med-VQA and report generation tasks, demonstrating superior performance.

\section*{Acknowledgement}
This work is partially supported by Cisco Faculty Research Award and NIH R01AG085581, R01AG079291 and P50HD103573. The Authors acknowledge the National Artificial Intelligence Research Resource (NAIRR) Pilot, NCSA DeltaAI and OpenAI API for contributing to this research result.


\section*{Impact Statement}

The broader impact of this work lies in its potential to enhance the reliability and accuracy of AI-driven medical diagnostics by reducing hallucinations and improving visual-textual alignment in Med-LVLMs. This advancement could lead to more trustworthy AI tools in healthcare, benefiting patient outcomes. However, ethical considerations are crucial to ensure responsible deployment, prevent misuse, and avoid over-reliance on AI-generated medical advice. Future societal benefits may include reduced diagnostic errors and improved healthcare efficiency, but ongoing research and ethical oversight are essential to align these advancements with the best interests of patients and providers.

\nocite{langley00}

\bibliography{example_paper}
\bibliographystyle{icml2025}

\newpage
\appendix
\onecolumn

\appendix

\section{Data}
\label{sec:data}

\subsection{Data Statistics}
The data statistics are shown in Table~\ref{tab:data_statistics_vqa} and Table~\ref{tab:test_set_statistics}. In the training datasets, the reported quantities for the two datasets in report generation represent image-report pairs, while the quantities for the two datasets in the medical VQA task represent question-answer pairs.

\begin{table}[htbp]
\centering

\caption{Data statistics for the training set of four datasets under two different task settings. “Train (visual)” refers to the number of visual-only preference data, while “Train (text)” indicates the number of text-only preference data.}

\vspace{0.5em}
\resizebox{0.5\linewidth}{!}{
\begin{tabular}{l|c|c|c}
\toprule
\textbf{Dataset} & \textbf{Train (visual)} & \textbf{Train (text)} & \textbf{Train (all)} 
\\
\midrule
IU-Xray & 2069 & 2069 & 4138  \\
MIMIC-CXR & 800 & 800 & 1600  \\ \midrule
SLAKE & 4919 & 4919 & 9838  \\
VQA-RAD & 1797 & 1797 & 3594  \\
\bottomrule
\end{tabular}
}
\label{tab:data_statistics_vqa}
\end{table}

\begin{table}[htbp]
\centering
\caption{Data statistics of test set. \#Images, \#QA items and \#Reports mean the number of images, QA pairs and reports, respectively.}
\footnotesize
\vspace{0.5em}
\begin{tabular}{l|c|c|c}
\toprule
\textbf{Dataset} & \textbf{\#Images} & \textbf{\#QA items} & \textbf{\#Reports} \\
\midrule
IU-Xray    & 590 & -    & 590 \\
MIMIC-CXR  & 200 & -    & 200 \\
\midrule
SLAKE      & 641 & 1061 & -   \\
VQA-RAD    & 315 & 451  & -   \\
\bottomrule
\end{tabular}
\label{tab:test_set_statistics}
\end{table}
\subsection{Involved Datasets}
We leverage four open-source medical vision-language datasets: MIMIC-CXR~\cite{johnson2020mimic}, IU-Xray~\cite{demner2016preparing}, SLAKE~\cite{liu2021slake}, and VQA-RAD~\cite{lau2018dataset}. These datasets are designed for different tasks: the first two focus on medical report generation, while the latter two are tailored for medical visual question answering.
\begin{itemize}
    \item \textbf{IU-Xray} is a dataset that includes chest X-ray images and corresponding diagnostic reports.
    \item \textbf{MIMIC-CXR} is a widely accessible dataset containing chest X-ray images in DICOM format along with corresponding radiology reports.
    \item \textbf{SLAKE} is an English-Chinese bilingual dataset comprising 642 images and 14,028 question-answer pairs designed for training and evaluating Med-VQA systems.
    \item \textbf{VQA-RAD} is the first dataset manually curated by clinicians, featuring naturally occurring questions about radiology images along with corresponding reference answers.
\end{itemize}

\section{Hyperparameter Settings}
\label{sec:setting}
For the usage of visual tools, we employ “disease” as the text description to guide MedKLIP~\cite{wu2023medklip} in generating heatmaps. 
For multi-agent collaboration, the process is conducted over 5 rounds
. During score normalization, the parameters are set as: $\alpha = 0.75$, $\beta = 1.25$, $\mu = 1$, and $\sigma^2 = 0.1$. All hyperparameters are kept consistent across the experiments to eliminate any potential bias introduced by hyperparameter tuning. All experiments are implemented using PyTorch 2.1.2 on four NVIDIA RTX A6000 GPUs, with training requiring approximately 2 to 3 hours.

\begin{table*}[t]
\centering
\footnotesize
\caption{Detailed performance comparison on report generation tasks covering IU-Xray and MIMIC-CXR datasets. BL denotes BLEU.}
\vspace{0.5em}
\resizebox{0.98\linewidth}{!}{
\begin{tabular}{l|cccccc|ccccccc}
\toprule
\multirow{2}{*}{Models} & \multicolumn{6}{c}{\textbf{IU-Xray}} & \multicolumn{6}{c}{\textbf{MIMIC-CXR}} \\ 
 & BL-1  & BL-2  & BL-3  & BL-4  & ROUGE-L & METEOR  & BL-1  & BL-2  & BL-3  & BL-4  & ROUGE-L & METEOR \\ \midrule
LLaVA-Med v1.5   	&38.42&	13.40&	4.74&	1.67& 10.31 & 10.95 & 29.41	&10.19	&3.58	&1.26&	9.38&	7.71\\ \midrule
+ Self-Rewarding & 38.25&	13.17	&3.61&	1.08 & 10.38 & 10.52  & 29.29&	10.32&	3.67	&1.30& 9.27 & 7.73\\

+ DPO      & 41.63&	15.13	&5.56&	2.03 & 12.95 & 17.13 & 29.61&	10.29	&3.61	&1.27&9.45&	7.81&\\ 
+ POVID & 50.84	&20.65&	8.38&	3.31 & 24.33 & 30.05 & 29.68&	10.29&	3.61&	1.26& 9.66 & 7.84 \\
+ SIMA & 42.67	&16.82	&5.98&	2.95 & 22.87 & 29.10 & 29.58&	10.23&	3.59&	1.24& 9.58 & 7.49 \\
+ FiSAO & 51.10	&20.92&	8.64	&3.59 & 25.72 & 30.82 & 29.76	&10.37&	3.74	&1.39 & 9.68 & 7.62 \\
+ STLLaVA-Med & 42.38	&15.27&	5.59&	1.20& 10.58 & 10.51 & 29.33	&10.27&	3.58	&1.27& 9.29 & 7.72\\
 + \textbf{\ours} (Ours) & 55.58&	23.93&	10.36&	4.40&29.52&	34.16 & 33.67&	11.91&	4.28&	1.54&	11.13&	10.03 \\
\bottomrule
\end{tabular}
}
\label{tab:detailed_bench}
\end{table*}

\begin{table*}[t]
\centering
\footnotesize
\caption{Detailed component ablation study on report generation tasks covering IU-Xray and MIMIC-CXR datasets. BL denotes BLEU. Here, stage 1 and stage 2 denotes generating hallucinated medical responses and adding noise to localized lesion regions, respectively. }
\vspace{0.5em}
\resizebox{\linewidth}{!}{
\begin{tabular}{l|cccccc|ccccccc}
\toprule
\multirow{2}{*}{Models} & \multicolumn{6}{c}{\textbf{IU-Xray}} & \multicolumn{6}{c}{\textbf{MIMIC-CXR}} \\ 
 & BL-1  & BL-2  & BL-3  & BL-4  & ROUGE-L & METEOR  & BL-1  & BL-2  & BL-3  & BL-4  & ROUGE-L & METEOR \\ \midrule
+ Stage 1 (Single-LLM)   & 43.45&	16.05&	5.99&	2.21 & 19.66 & 22.65 & 29.41&	10.19&	3.58&	1.26&	9.33&	7.77\\ 
+ Stage 1 (Multi-LLMs) & 43.95&	16.44&	6.21&	2.31      & 19.57       & 22.92    & 29.85	&10.38	&3.65	&1.28& 9.62 & 8.18\\ 
+ Stage 2 & 55.15&	23.59&10.13&	4.23&	29.02&	34.26& 30.96&	10.89&3.87&	1.38&	9.85&	8.81\\
+ Stage 1+2 (Single-LLM)& 55.36&	23.85&	10.34&	4.39&	29.30&	34.22& 32.96	&11.63&	4.14&	1.46&	10.99 &10.03\\
+ Stage 1+2 (Multi-LLMs)& 55.58&	23.93&	10.36&	4.40&	29.52&	34.16 & 33.67&	11.91&	4.28&	1.54&	11.13 &10.05 \\
\bottomrule
\end{tabular}
}
\label{tab:ablation_appendix}
\end{table*}

\section{Involved Baselines}
\label{sec:baseline}
\begin{itemize}
\item\textbf{DPO}~\cite{rafailov2023direct} is a fine-tuning approach designed to align large language models (LLMs) with human preferences in a stable, efficient, and computationally lightweight manner. Unlike traditional Reinforcement Learning from Human Feedback (RLHF), which involves training a reward model and using reinforcement learning to maximize the reward, DPO simplifies the process by reframing the problem. It parameterizes the reward model in a way that allows the optimal policy to be derived directly through a classification loss, eliminating the need for complex sampling or extensive hyperparameter tuning during fine-tuning.
\item\textbf{Self-Rewarding}~\cite{yuanself} is a novel approach where the language model itself acts as a judge, generating rewards via LLM-as-a-Judge prompting during training. Unlike traditional methods that rely on reward models trained from human preferences, which are limited by human performance and static design, this method enables the model to iteratively improve both its instruction-following abilities and its reward-generating quality during iterative DPO training.

\item\textbf{STLLaVA-Med}~\cite{sun2024stllava} refines the preference selection process using GPT-4o and applies it in medical vision-language tasks. STLLaVA-Med extends the DPO approach by incorporating a self-training mechanism specifically tailored for the medical domain.

\item \textbf{POVID}~\cite{zhou2024aligning} addresses the hallucination problem in vision-language models by generating feedback data using AI models. It uses ground-truth instructions as preferred responses and creates dispreferred data by injecting plausible hallucinations and distorting images, integrating these strategies into an RLHF pipeline via DPO.

\item \textbf{FiSAO}~\cite{cui2024fine} introduces a fine-grained self-alignment optimization method that leverages the model's own visual encoder to improve vision-language alignment. By utilizing token-level feedback from the vision encoder, it enhances alignment without the need for additional external data, outperforming traditional preference tuning methods.

\item \textbf{SIMA}~\cite{wang2024enhancing} is a framework that enhances visual and language modality alignment through self-improvement, eliminating the need for external models or data. It uses prompts from existing datasets to self-generate responses and employs an in-context self-critic mechanism with vision metrics to select optimal response pairs for preference tuning.

\end{itemize}

\section{Additional Results}

In this section, we present a detailed benchmark analysis for the report generation task. Table~\ref{tab:detailed_bench} compares our method with other baseline approaches. Additionally, Tables~\ref{tab:ablation_appendix} and \ref{tab:performance_comparison} provide comprehensive component ablation results for both the Medical VQA and report generation tasks.

\begin{table}[htbp]
\centering
\caption{Detailed component ablation study on SLAKE and VQA-RAD datasets for both open and closed settings. Here, stage 1 and stage 2 denotes generating hallucinated medical responses and adding noise to localized lesion regions, respectively. }
\vspace{0.5em}
\footnotesize
\begin{tabular}{l|cc|cc}
\toprule
\textbf{Method} & \multicolumn{2}{c|}{\textbf{SLAKE}} & \multicolumn{2}{c}{\textbf{VQA-RAD}} \\
 & Open & Close & Open & Close \\
\midrule
Stage 1 (Single-LLM) & 47.99 & 64.18 & 32.27 & 65.07 \\
Stage 1 (Multi-LLMs) & 49.39 & 65.87 & 32.42 & 69.85 \\ 
Stage 2 & 51.25 & 68.51 & 31.09 & 62.87 \\
\bottomrule
\end{tabular}
\label{tab:performance_comparison}
\end{table}

\section{Prompts}
\begin{figure}[htbp]
    \centering
    \includegraphics[width=0.6\linewidth]{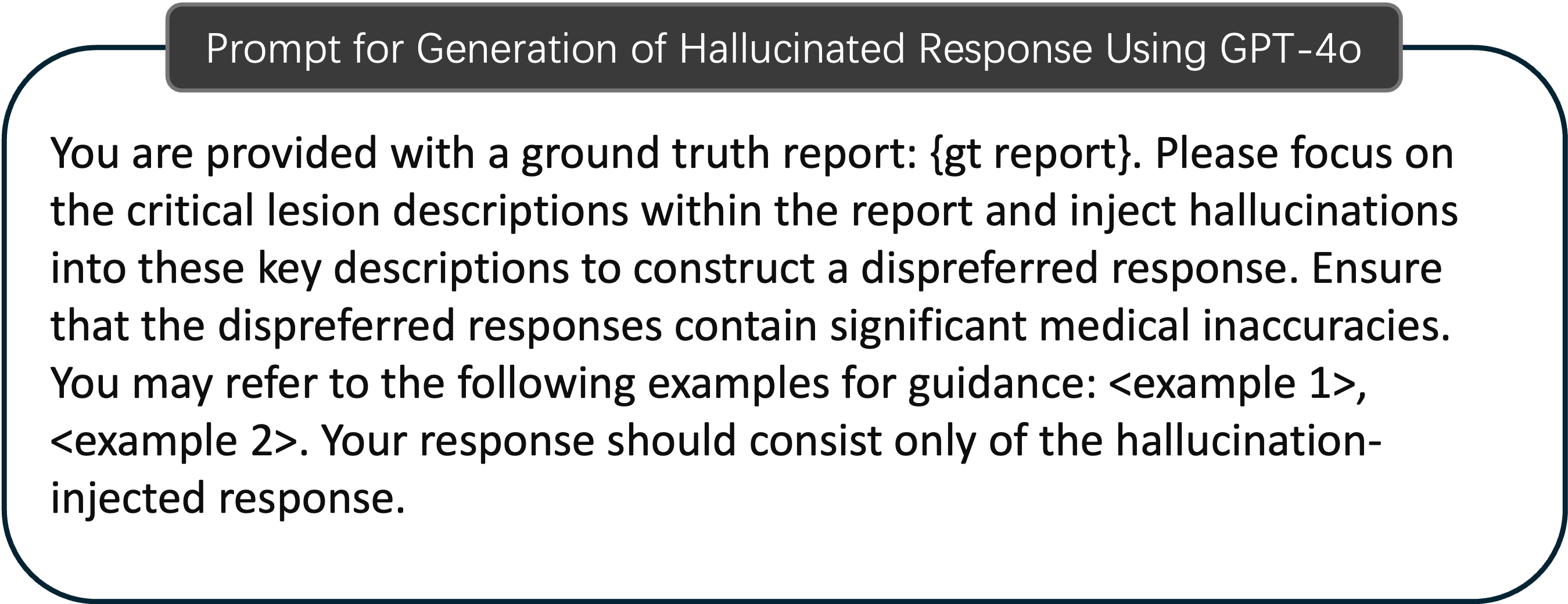}  
    \caption{The instruction to GPT-4o for the  rejected hallucinated answer. }
    \label{fig:gpt_prompt}
\end{figure}

\begin{figure}[htbp]
    \centering
    \includegraphics[width=0.6\linewidth]{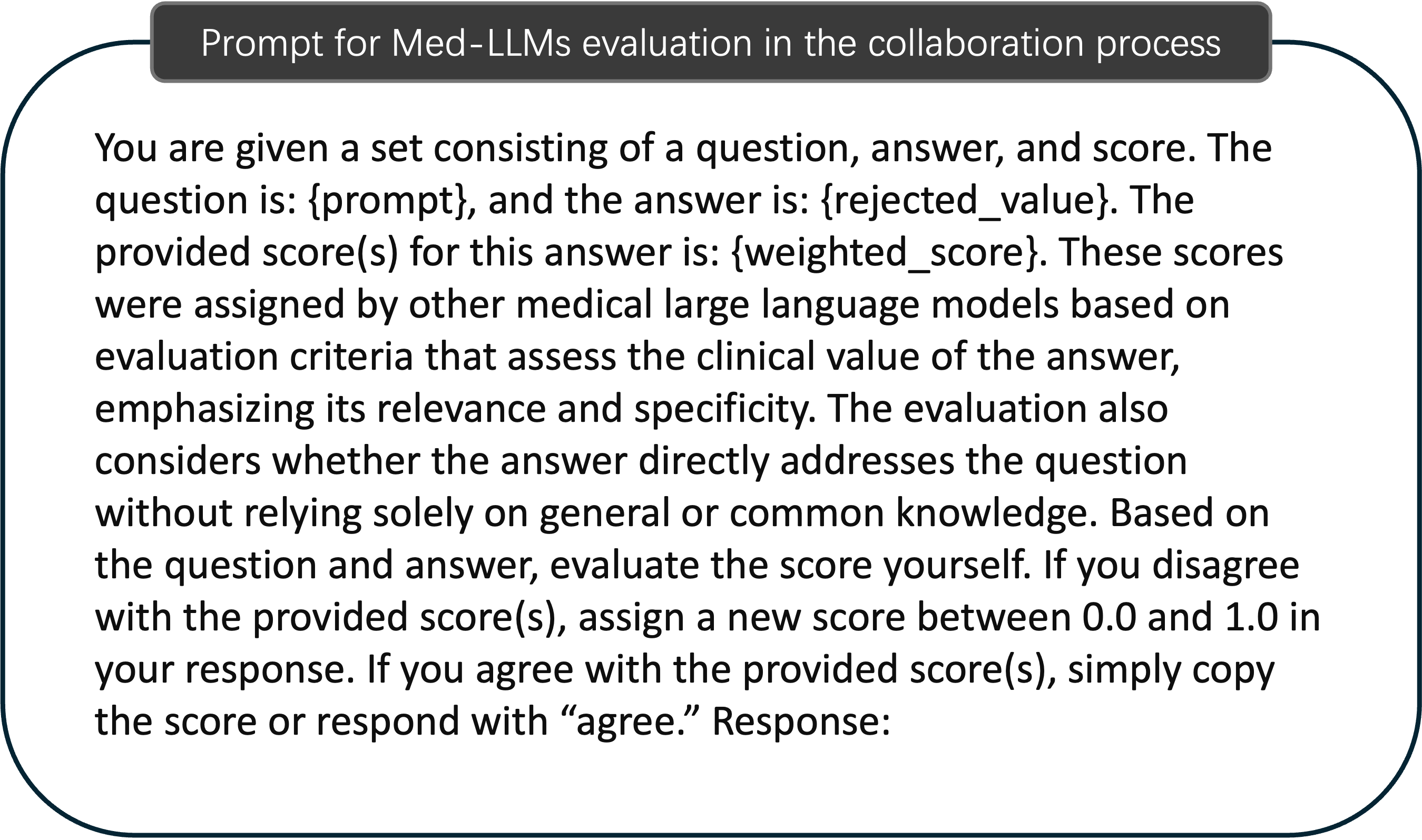}  
    \caption{The instruction to Med-LLMs for evaluating and generating clinical relevance score. }
    \label{fig:med-llm_eval_prompt}
\end{figure}

We utilize GPT-4o to generate hallucinated responses for constructing preference data, as illustrated by the prompts in Figure~\ref{fig:gpt_prompt}. Subsequently, a multi-agent system comprising Med-LLMs is employed to evaluate the clinical relevance scores of these rejected responses, with the evaluation prompts shown in Figure~\ref{fig:med-llm_eval_prompt}.

%

\section{More Cases}

\begin{figure*}[h]
  \centering
  \includegraphics[width=1.0\textwidth]{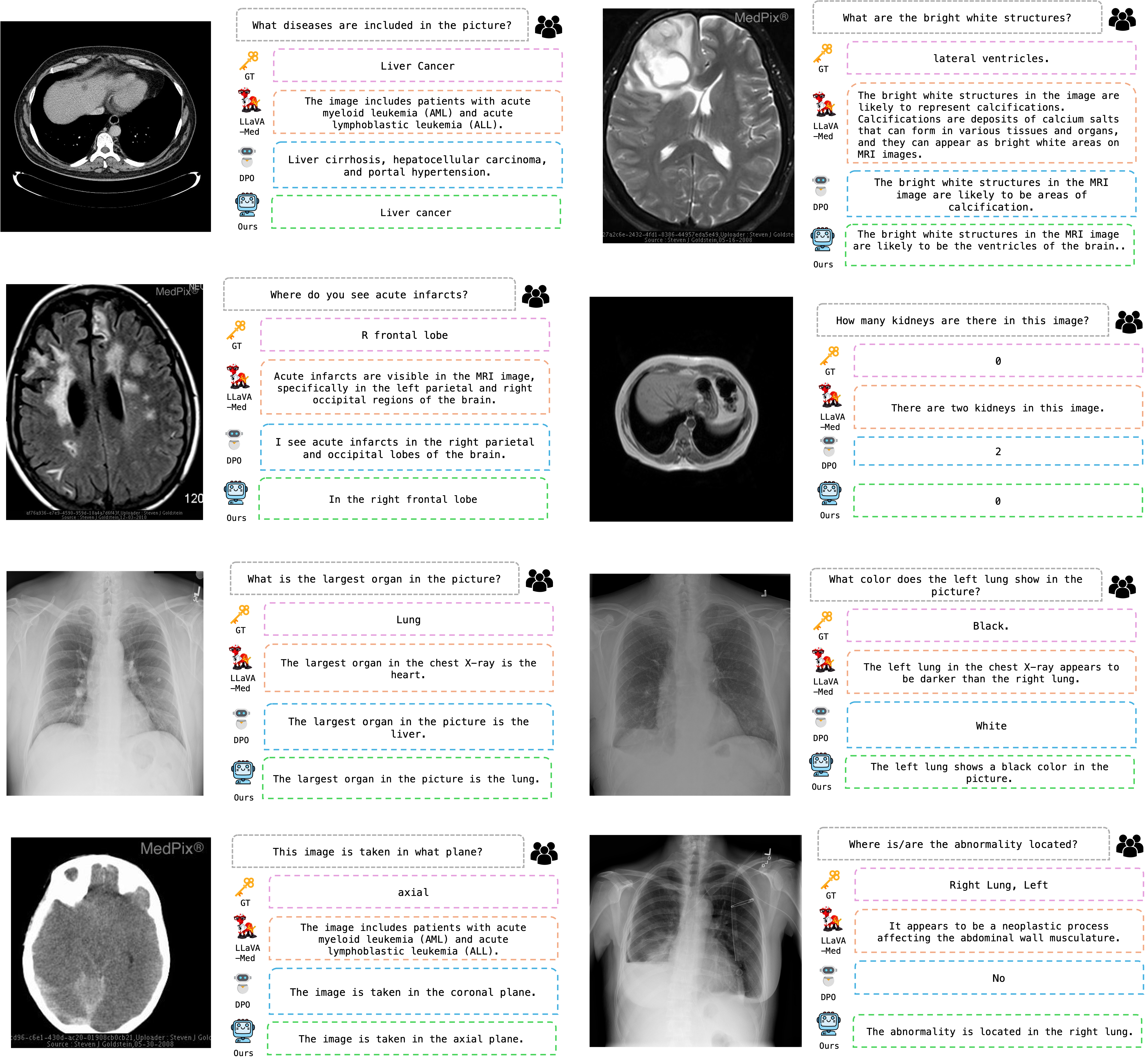}  
\caption{More cases that reduce hallucinated errors.}
  \label{fig:all_case}
  \vspace{0.5em}
\end{figure*}

We present additional examples in Figure~\ref{fig:all_case}, illustrating how our method effectively reduces hallucinated errors.

\end{document}